%% file: main.tex
\documentclass[conference]{IEEEtran}

\IEEEoverridecommandlockouts

\usepackage{cite}
\usepackage{amsmath,amssymb,amsfonts}
\usepackage{graphicx}
\usepackage{textcomp}
\usepackage{xcolor}

\usepackage{makecell} 
\usepackage{nicematrix, tikz} 
\usepackage{subcaption}
\usepackage{booktabs} 
\usepackage{algorithm} 
\usepackage{algpseudocode} 
\usepackage[dvipsnames]{xcolor} 

\def\BibTeX{{\rm B\kern-.05em{\sc i\kern-.025em b}\kern-.08em
    T\kern-.1667em\lower.7ex\hbox{E}\kern-.125emX}}
\begin{document}

\bstctlcite{IEEEexample:BSTcontrol} 

\title{\LARGE \textbf{FedCKA: Representation-Guided Layer Personalization for Federated 3D Perception Across Driving Domains}

\thanks{\textsuperscript{1}Jolle Verhoog, \textsuperscript{1}Ali Burak Ünal, and \textsuperscript{1}Holger Caesar are with the faculty of Cognitive Robotics, Mechanical Engineering, TU Delft, Delft, Netherlands. \tt\small[jtverhoog, a.b.unal, h.caesar]@tudelft.nl}
}

\author{\IEEEauthorblockN{Jolle Verhoog\textsuperscript{1}, Ali Burak Ünal\textsuperscript{1}, and Holger Caesar\textsuperscript{1}}}

\maketitle


\enlargethispage{4\baselineskip} 

\input{sec/0_abstract}    
\input{sec/1_introduction}
\input{sec/2_related_works}
\input{sec/3_proposed_method}
\input{sec/4_experiment_design}

\input{sec/5_experiments}
\input{sec/6_discussion}
\input{sec/7_limitations}
\input{sec/8_conclusion}
\input{sec/9_acknowledgements}

\bibliographystyle{IEEEtran}
\bibliography{main}


\end{document}

%% file: sec/0_abstract.tex
\begin{abstract}
Robust perception in intelligent vehicles demands 3D object detectors that remain dependable under domain shifts, such as changes in time of day, location, or weather. However, due to costly annotation and rare shifts, some environments lack sufficient data to train a standalone detector. Federated learning offers a privacy-preserving framework for collaborative model training, enabling clients to benefit from shared learning across diverse environments. Yet, this framework traditionally relies on a single global consensus model, which struggles to perform across heterogeneous local data distributions. Local conditions are better captured by adapting a subset of the model, but many personalization approaches rely on predefined layer partitions or fixed personalization ratios, thereby limiting adaptation to client-specific divergence.
To reduce this rigidity, we propose FedCKA, a Centered Kernel Alignment (CKA)-based strategy that dynamically handles the personalization-globalization trade-off. Specifically, FedCKA computes layer-wise feature similarities between local client models and the global consensus model during training. By converting layer-wise similarity scores into client-specific aggregation masks, FedCKA selectively shares representation-consistent layers.
Evaluation on a unified multi-domain benchmark based on nuScenes shows that FedCKA outperforms established federated baselines, including FedBN, FedRep, and FedSelect, improving average NDS by 7 percentage points over the strongest baseline. The findings offer both a comparative benchmark and a promising direction for robust federated 3D perception across shifts in location, weather, and illumination. Code is available at \color{blue}https://github.com/j-verhoog/FedCKA\color{black}.
\end{abstract}

%% file: sec/1_introduction.tex
\section{Introduction} \label{sec:introduction}
\begin{figure}[t]
    \centering
    \resizebox{0.75\linewidth}{!}{%
        \begin{minipage}{\linewidth}
            \begin{subfigure}[t]{0.5\linewidth}
                \includegraphics[width=\linewidth]{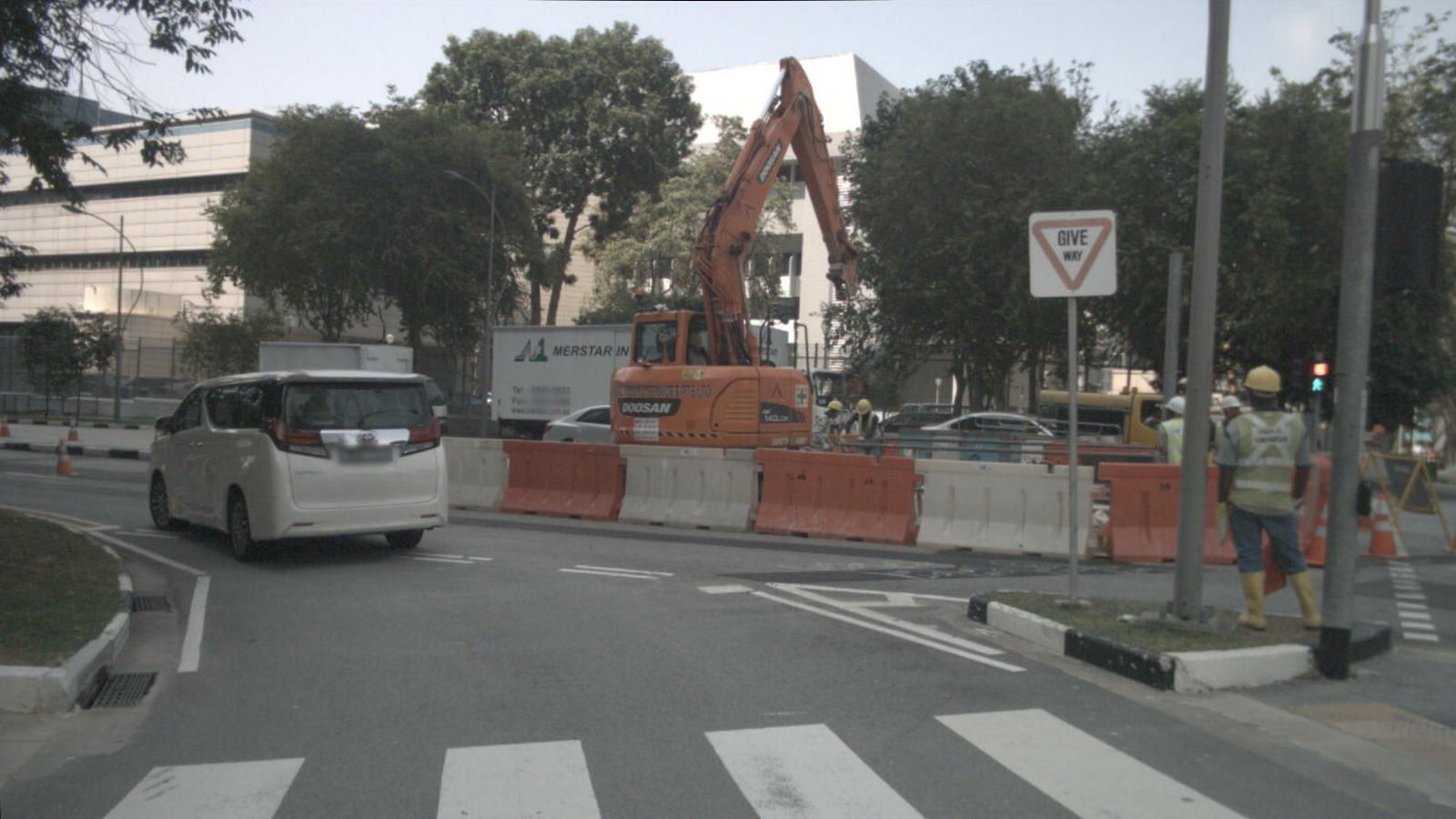}
            \end{subfigure}%
            \begin{subfigure}[t]{0.5\linewidth}
                \includegraphics[width=\linewidth]{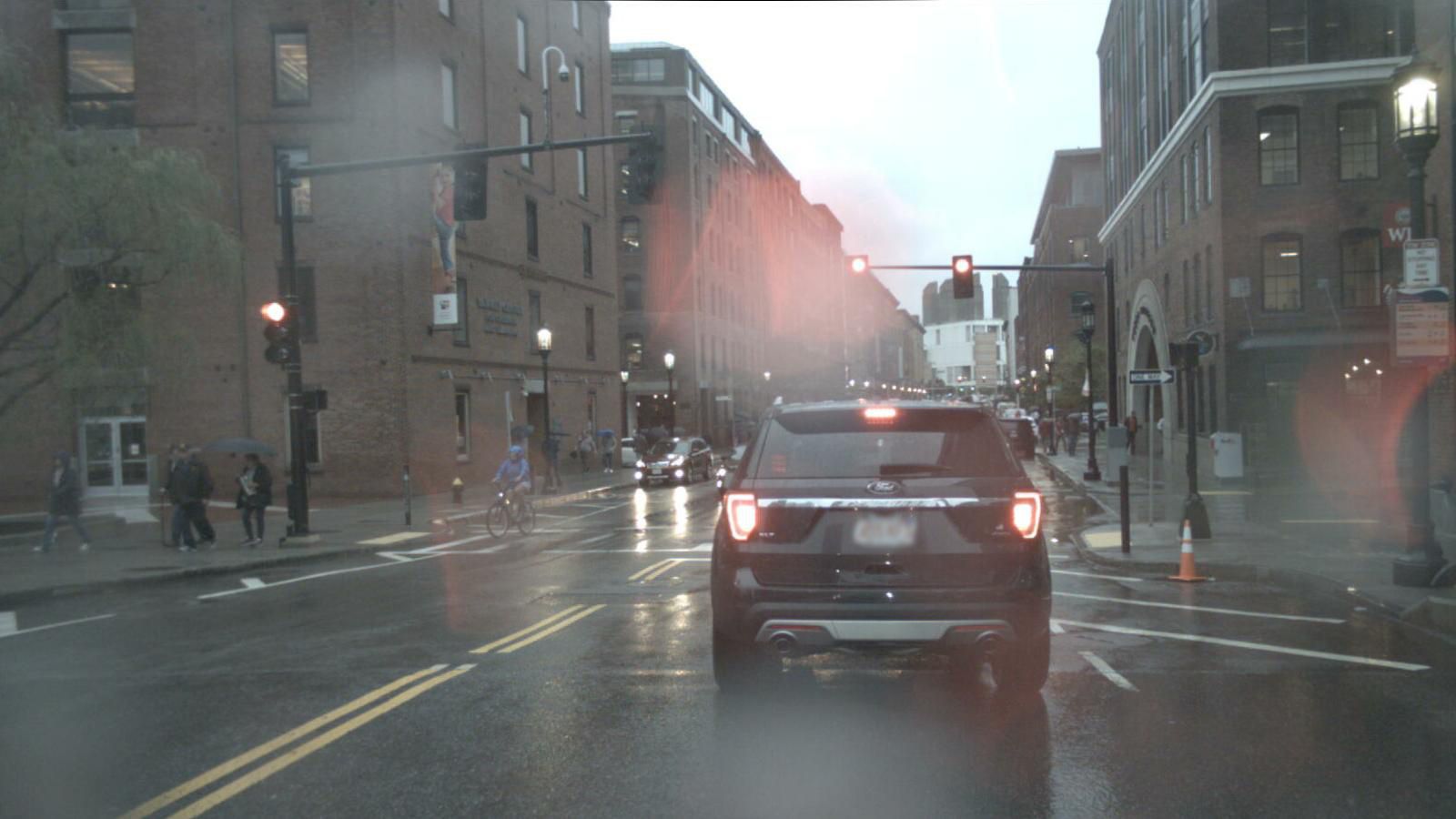}
            \end{subfigure}\\[-0.5em]
            \begin{subfigure}[t]{0.5\linewidth}
                \includegraphics[width=\linewidth]{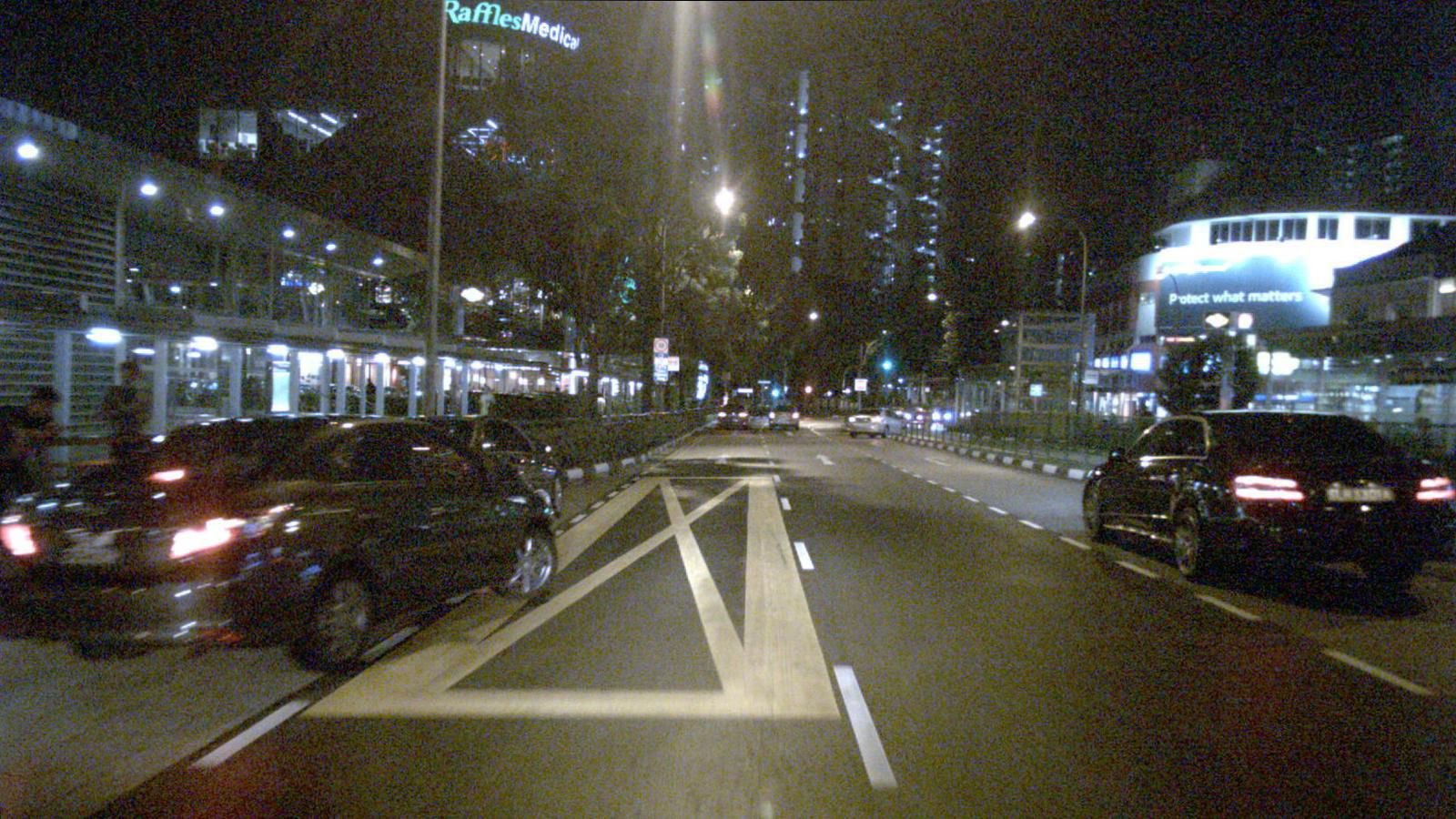}
            \end{subfigure}%
            \begin{subfigure}[t]{0.5\linewidth}
                \includegraphics[width=\linewidth]{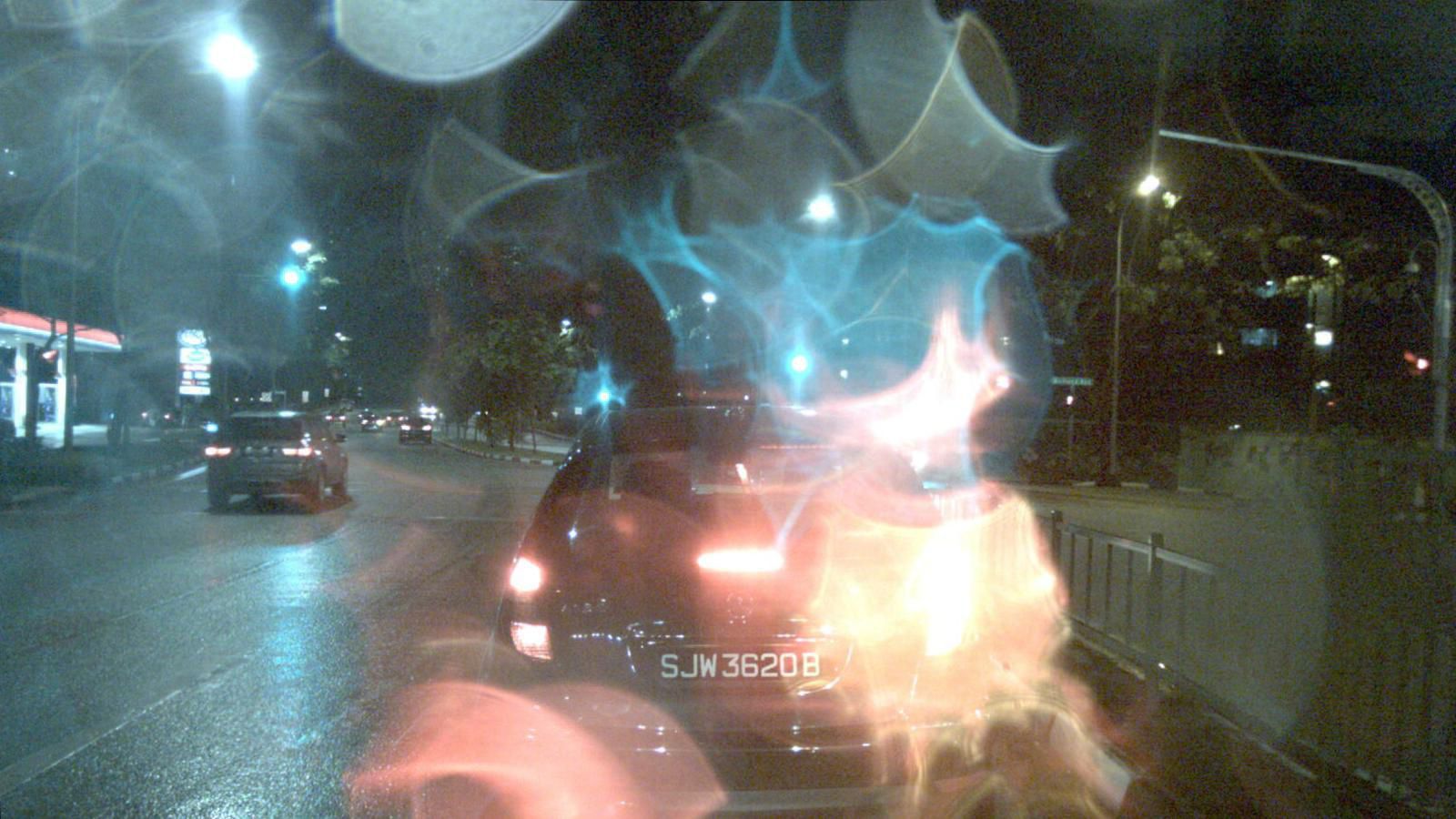}
            \end{subfigure}
        \end{minipage}%
    }
    \caption{nuScenes~\cite{Caesar2020nuScenes} samples showing clear versus rainy weather (columns) across Boston daytime and Singapore nighttime conditions (rows).}
    \label{fig:nuscenesexampl}
    \vspace{-1em}
\end{figure}

3D Object Detection (3D-OD) is central to robotic perception in Intelligent Vehicles (IV), providing object poses and dimensions needed for planning and safe navigation.
However, domain shifts across vehicles, times of day, weather, and locations substantially degrade 3D detection performance~\cite{nuScenesC} (see Fig.~\ref{fig:nuscenesexampl}).
Because labeled data in niche or edge-case domains can be too scarce to support robust standalone training, a collaborative framework is desirable. 
Yet, for IV applications, centrally collecting multimodal sensor data is constrained by privacy barriers and massive data volumes~\cite{mcmahan2017communicationefficientlearningdeepnetworks}, while naive localized retraining remains computationally expensive~\cite{strubell-etal-2019-energy}.

These constraints motivate Federated Learning (FL), which enables privacy-preserving collaborative training for multi-domain adaptation without data centralization~\cite{li2021fedbn, li2020fedprox, karimireddy2020scaffold}, by sharing only model updates. However, real-world deployment across heterogeneous data distributions introduces a central personalization-globalization trade-off: which model components should be globally shared versus locally personalized? Many personalization methods resolve this trade-off through static layer partitions or fixed personalization ratios, which may fail to capture client- and layer-wise representation divergence.
For multi-modal 3D-OD, this rigidity is particularly problematic because domain shifts can affect sensor backbones and fusion modules asymmetrically~\cite{yosinski2014transferable_diffpartsdifffeatures}.

Despite this need, FL for multimodal 3D-OD remains underexplored. While prior federated detection and, more recently, federated 3D-OD studies exist, they do not systematically isolate location-, weather-, and illumination-induced client heterogeneity. 
This limits insight into whether personalization helps under these structured client-level domain shifts.
While recent federated 3D detection approaches, including FedM2Former~\cite{fedm2former} and FedPillarNet~\cite{fedpillarnet}, address client heterogeneity through personalized learning, they do not systematically isolate location-, weather-, and illumination-induced domains or compare representative FL personalization strategies within a unified 3D-OD benchmark.

To address this gap, we study how federated 3D-OD methods handle location-, weather-, and illumination-induced client heterogeneity, and whether representation consistency can determine which layers should be globally aggregated or locally personalized.
We isolate client-data heterogeneity by focusing on labeled, client-level domain adaptation across location, weather, and illumination, leaving system-level constraints such as client reliability and sensor-configuration variability to future work.
We make the following contributions:

\begin{enumerate}
\item Establish a unified multi-domain federated 3D object detection benchmark based on explicit location-, weather-, and illumination-induced client heterogeneity, enabling controlled evaluation under realistic domain shifts.
\item Introduce FedCKA, a dynamic CKA-based layer masking strategy that uses layer-wise representation similarity to determine which model components should be globally aggregated versus locally personalized.
\item Demonstrate that FedCKA outperforms established federated baselines, including FedBN, FedRep, and FedSelect, improving average NDS by 7 percentage points over the strongest competitor by identifying domain-sensitive layers.
\end{enumerate}

%% file: sec/2_related_works.tex
\section{Related Work} \label{sec:relatedworks}

\textbf{3D Object Detection} methods are commonly categorized by sensor modality: camera, LiDAR, or multimodal~\cite{s25092794}.
While multimodal methods typically fuse LiDAR and camera data via Bird's-Eye-View (BEV) representations~\cite{Cai_Wu_2025}, they struggle across varying domains like day, night, and adverse weather~\cite{MultiCorrupt}. 
This limitation motivates methods that can exploit multi-domain data, particularly when such data are distributed across clients and cannot be centrally pooled.

\textbf{Federated Learning} (FL) enables such collaborative training on decentralized data. It commonly operates under client data that is not independent and identically distributed (IID), not centrally shared, or not of equal size across clients~\cite{zhan}.
This statistical and system heterogeneity introduces challenges for convergence~\cite{li2020fedprox} and cross-domain generalization~\cite{li2025federateddg_crossdomaingeneralization}.

The foundational FedAvg~\cite{mcmahan2017communicationefficientlearningdeepnetworks} algorithm aggregates independent local updates via weighted averaging. While effective under IID conditions, its performance degrades significantly under heterogeneous data distributions~\cite{hsu2020federated_noniiddegradation}.

\textbf{Data heterogeneity in FL} fundamentally shapes what client models should share and how the server should aggregate their updates. Approaches to managing this heterogeneity span data preprocessing, training objectives, and aggregation schemes~\cite{zhan}. 
For all of these, the main distinction is the objective: aligning all clients toward one domain-invariant model, or producing multiple domain-adapted models.

To learn a more domain-invariant global model, methods such as FedProx~\cite{li2020fedprox}, FedDyn~\cite{acar2021feddyn}, and SCAFFOLD~\cite{karimireddy2020scaffold} reduce client drift during training.
Other approaches use task arithmetic~\cite{ilharco2023task_arithmetic} to align client update gradients and estimate a less conflicting global update direction~\cite{yu2020pcgrad,zhang2024fedgh, nguyen2025fedomg}.
However, a single global model may be suboptimal under strong client heterogeneity, and these methods require problem-specific manual parameter tuning. 
Personalized FL addresses this by adapting models to client-specific domains while retaining shared knowledge.
Common approaches keep batch normalization or other targeted layers client-specific, as in FedBN~\cite{li2021fedbn} and FedRep~\cite{collins2021fedrep}.
Because these methods require manually predefined global and domain-specific layer partitions, this static approach can be suboptimal when domain specificity varies across layers or clients.

Recent work addresses this rigidity by using model updates to infer which parameters to mask for personalization.
FedSelect~\cite{tamirisa2024fedselect} assumes the largest parameter updates are the most domain-significant and personalizes them.
FedMC~\cite{fedmc} uses the Fisher information derived from gradient magnitudes to identify the most sensitive parameters for personalization.
Although these methods avoid manual partitioning, they still rely on parameter-space importance by implicitly assuming that same-index parameters across clients correspond to comparable functional components, which may not hold~\cite{singh2020model}. To overcome the limitations of parameter-space comparisons and manual layer selection, this work selects personalized layers based on representational consistency across clients.

%% file: sec/3_proposed_method.tex
\begin{figure*}[t]
\centering
\includegraphics[width=\textwidth]{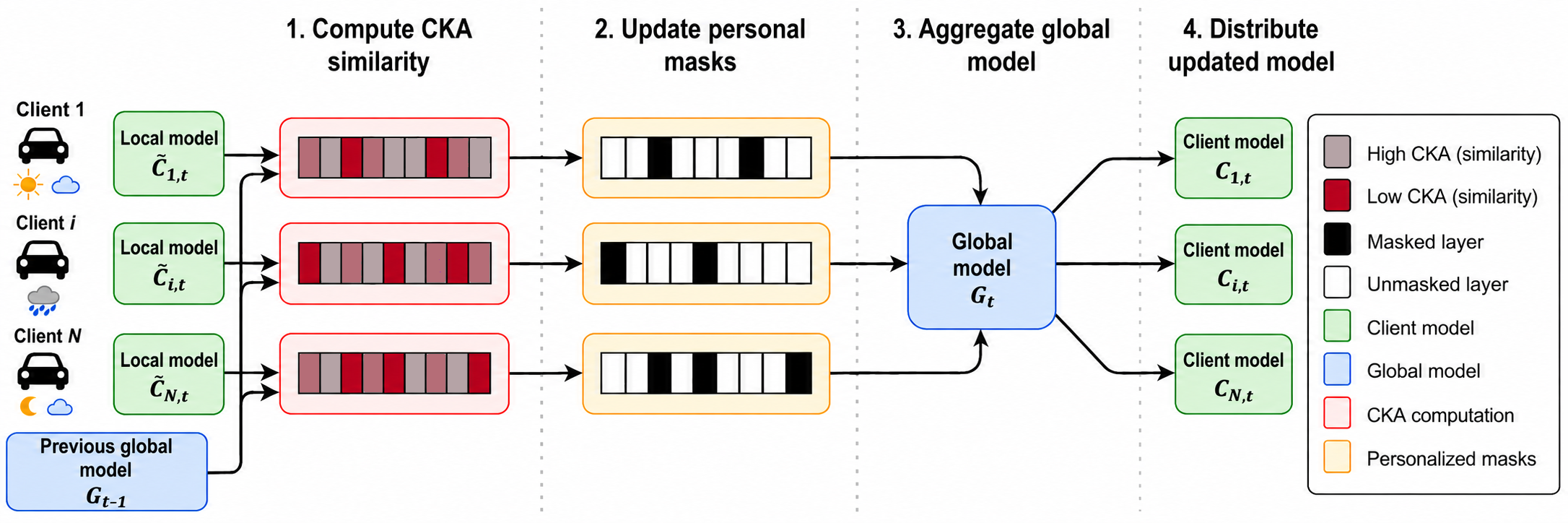}
\caption{
Proposed client-specific aggregation via CKA-based layer masking. At round $t$, each client compares its locally updated model $\widetilde{C}_{i,t}$ with the previous global model $G_{t-1}$ using layer-wise CKA to generate a binary mask. Only unmasked layers are transmitted and aggregated by the server to form $G_t$. $G_t$ is returned to the clients, where each client overwrites its unmasked layers while masked layers retain their local parameters to obtain $C_{i,t}$.
}
\label{fig:proposed_method}
\end{figure*}

\section{Proposed Method} \label{sec:method}
\subsection{Motivation}
FedCKA is motivated by the idea that layers should be shared across clients only when their representations remain consistent after local adaptation.
While existing federated methods rely on parameter-space distances~\cite{tamirisa2024fedselect} or gradient magnitudes~\cite{fedmc}, this work argues that representational consistency is a more reliable proxy for transferability and uses it to determine which layers to personalize. 
This strategy incorporates Centered Kernel Alignment (CKA)~\cite{kornblith2019similarity_CKA} to measure how differently two layers map the same set of incoming features to their outgoing features. 
This metric produces a normalized score bounded in $[0,1]$, with 0 indicating strong dissimilarity and 1 indicating high similarity in the representation space transformation for each layer. 

By comparing layers at identical depths across the global and local models, FedCKA estimates whether each layer remains transferable. 
The most dissimilar layers can then be personalized by masking them from the global model aggregation. This lets certain layers drift to domain-specific optima, while sharing representationally similar layers to transfer global knowledge.
Because only unmasked parameters are transmitted, FedCKA reduces shared parameter information by the fraction of masked parameters, improving bandwidth efficiency and potentially reducing information exposure.

\subsection{FedCKA System Architecture}
As illustrated in Fig. \ref{fig:proposed_method}, FedCKA entails coordinated exchanges between the clients and the central server at every communication round.
Let $i\in{1,\ldots,N}$ denote the client index and $t$ the communication round, with $G_t$ denoting the global model after round $t$ and $C_{i,t}$ the corresponding client model. During round $t$, client $i$ locally trains $C_{i,t-1}$ for $E$ local training epochs, yielding the locally updated model $\widetilde{C}_{i,t}$.

The first step is to estimate local divergence by computing the CKA score between the saved state of the previous global model, $G_{t-1}$, and the newly updated local model $\widetilde{C}_{i,t}$.
Using $G_{t-1}$ as a common reference ensures that each client's layer-wise divergence is measured against the same pre-aggregation model state.
This computation uses a small random subset of the client's local training data passed through both models without requiring ground-truth labels.
Based on the layers' representational similarity, the client determines which layers to personalize and thus mask from global aggregation.
Consequently, each client sends its unmasked layers, alongside a lightweight binary mask identifying them, to the central server.

The server aggregates these shared parameters layer-by-layer to form the updated global model $G_t$.
Finally, each client updates $\widetilde{C}_{i,t}$ by injecting the unmasked parameters from $G_t$, yielding the updated client model $C_{i,t}$.
This process allows each client to personalize its most dissimilar layers while benefiting from the knowledge learned by other clients.

\subsection{Centered Kernel Alignment}
Centered Kernel Alignment~\cite{kornblith2019similarity_CKA} is a similarity index between two layers, based on the Hilbert-Schmidt Independence Criterion (HSIC)~\cite{gretton2005measuring_HSIC} for comparing the representations induced by two layers. 
Different layer parameterizations may yield similar representations, making direct comparison in parameter space misleading. 
CKA addresses this limitation by comparing layer activations rather than parameters. 
CKA has similarly been used in representation analysis of vision architectures for transformers and convolutional neural networks~\cite{raghu2021visiontransformers}.
In this work, this principle is used in FL, where CKA measures whether a locally trained client layer remains representationally aligned with the previous global model.

Given layer outputs $X \in \mathbb{R}^{n \times p_1}$ and $Y \in \mathbb{R}^{n \times p_2}$, where $n$ denotes the number of aligned activation samples and $p_1$ and $p_2$ the respective feature dimensions, their representational similarity is computed using CKA, which normalizes the empirical HSIC between their respective Gram matrices $K = XX^{T}$ and $L = YY^{T}$:
\begin{equation}
\operatorname{CKA}(X,Y) = \frac{\operatorname{HSIC}(K,L)}{\sqrt{\operatorname{HSIC}(K,K)\operatorname{HSIC}(L,L)}}.
\label{eq:cka_normalized}
\end{equation}

CKA, therefore, indicates whether the same samples produce a similar relational structure at a given layer.
Mean-centering and normalization make it invariant to isotropic scaling and orthogonal transformations.
Because CKA does not rely on one-to-one alignment between individual neurons or feature channels, the exact ordering of neurons does not affect the comparison. It therefore captures whether a layer similarly organizes the same input samples across representations, rather than whether individual activations match directly.

\subsection{Personalization Strategies} \label{subsec:personalization_strats_method}
At each communication round, the personalization masks are constructed with the layer-wise CKA scores using one of two strategies: selecting a target percentage of parameters, analogous to FedSelect~\cite{tamirisa2024fedselect} and FedMC~\cite{fedmc}, or selecting all layers below a CKA similarity threshold.

To formally define the personalization strategies, let $N$ be the total number of clients, each with $n_i$ local data samples. For client $i$ at round $t$, let $w_{i,l}^{t}$ denote the weights of layer $l$ and $s_{i,l}^{t} \in [0,1]$ its computed CKA similarity score. 
This score is used to define the binary mask $M_{i,l}^{t} \in \{0,1\}$, where $M_{i,l}^{t}=1$ denotes personalization.
For each layer, the server aggregates the weight matrices only across the clients that keep that layer shared:

\begin{equation}
\bar{w}_{l}^{t} =
\frac{
\sum_{i=1}^{N} n_i(1-M_{i,l}^{t})w_{i,l}^{t}
}{
\sum_{i=1}^{N} n_i(1-M_{i,l}^{t})
}.
\end{equation}
If all clients personalize layer $l$, the denominator is zero, and the server keeps the previous global layer, i.e., $\bar{w}_{l}^{t}=\bar{w}_{l}^{t-1}$. 
Subsequently, each client $i \in \{1, \dots, N\}$ overwrites its local model by computing the following update for every layer:
\begin{equation}
\hat{w}_{i,l}^{t} = M_{i,l}^{t}w_{i,l}^{t} + (1-M_{i,l}^{t})\bar{w}_{l}^{t}, \quad \forall l  
\end{equation}

The persistence of the personalization mask across communication rounds differs between the two strategies.
For readability, client and round indices are omitted in the mask construction below. Depending on the chosen strategy, the mask \( M_l \) is constructed in one of two ways:

\textbf{Percentage-based masking:} 
For percentage-based masking, personalization is cumulative across communication rounds: once selected, a layer remains personalized for that client. We retain this permanent decoupling to align the masking protocol with FedSelect~\cite{tamirisa2024fedselect} and FedMC~\cite{fedmc}, enabling a more direct comparison in which the primary difference is the criterion used to identify parameters for personalization.

Let the layers be sorted by their CKA scores, denoted as $s_{(1)} \leq s_{(2)} \leq \dots \leq s_{(L)}$, where the subscript $(k)$ represents the layer index with the $k$-th lowest similarity score, $p_l$ be the parameter count of layer $l$, and $P$ be the total parameters in the network. Given a target personalization ratio $\rho \in [0, 1]$ at each communication round, we iteratively select the most dissimilar layers until their combined parameter count reaches the threshold $\rho P$. 
Specifically, we select the first $k$ sorted layers such that:
\begin{equation}
\sum_{j=1}^{k-1} p_{(j)} < \rho P \leq \sum_{j=1}^{k} p_{(j)}
\label{eq:percentage_masking_condition}
\end{equation}

The layers selected in the current round are then defined as $V = \{(1), \dots, (k)\}$, and the cumulative mask is applied as:
\begin{equation}
M_l =
\begin{cases}
1 & \text{if } M_l^{\mathrm{prev}} = 1 \text{ or } l \in V \\
0 & \text{otherwise}
\end{cases}
\label{eq:percentage_masking}
\end{equation}

Here, $M_l^{\mathrm{prev}}$ denotes whether layer $l$ was personalized in the previous communication round.
Since masking is performed at the layer granularity, the selected personalized ratio can exceed $\rho$ when the final selected layer, $(k)$, crosses the target.

\textbf{Threshold-based masking:}  Instead of enforcing a target proportion of the network to be personalized, this strategy evaluates the divergence of each layer independently. For a predefined similarity threshold $\tau \in [0, 1]$, a layer is personalized only if its CKA score indicates sufficient divergence from the global model:
\begin{equation}
    M_l = \begin{cases} 1 & \text{if } s_l < \tau \\ 0 & \text{otherwise} \end{cases}
    \label{eq:threshold_masking}
\end{equation}
Unlike the percentage-based approach, which dictates a fixed personalization footprint for each client regardless of actual layer alignment, threshold-based masking dynamically adjusts the volume of personalized parameters. 
Moreover, the threshold-based mask is not cumulative. 
Consequently, a layer personalized in round \(t-1\) can become globally shared again in round \(t\) if its CKA similarity no longer falls below \(\tau\).
Consequently, the fraction of parameters withheld can vary across clients and across rounds according to their layer-wise divergence, allowing the system to adapt.

%% file: sec/4_experiment_design.tex
\section{Experiment Design} \label{sec:exp_design}
\subsection{Dataset}
The experimental framework assigns each client to a distinct environmental domain. Table~\ref{tab:domain_scene_counts} defines these five domains (A-E), spanning Boston and Singapore across varying weather and lighting conditions.
This benchmark uses the nuScenes dataset, one of the most impactful datasets for onboard perception~\cite{10509812}, which distinguishes between time of day and weather.
The scenes, each consisting of roughly 40 annotated samples in a 20-second time window, are split such that each client acts as a distinct domain and possesses all data for that domain, as shown in Table~\ref{tab:domain_scene_counts}.
These clients exhibit heterogeneity not only in location, time of day, and weather, but also in scene count and class composition.
To illustrate, the Singapore training set contains, on average, three times as many annotated motorcycles, while Boston features more than twice as many cars. 
Although domain E's three validation scenes may yield noisy standalone metrics, their compounded domain shifts and the small training set provide a stress test for knowledge transfer at the edge of the federated system.

\begin{table}[t]
\centering
\caption{
Federated domain definitions and corresponding train/validation scene counts. 
``Sing.'' denotes Singapore.
}
\label{tab:domain_scene_counts}
\small
\setlength{\tabcolsep}{4pt}
\begin{tabular}{lccccc}
\toprule
Domain & A & B & C & D & E \\
\midrule
Location & Boston & Boston & Sing. & Sing. & Sing. \\
Time     & Day    & Day    & Day   & Night & Night \\
Weather  & Clear  & Rain   & Clear & Clear & Rain  \\
\midrule
Train scenes & 265 & 125 & 226 & 71 & 13 \\
Val scenes   & 53  & 24  & 58  & 12 & 3  \\
\bottomrule
\end{tabular}
\end{table}

\subsection{Detector}
To isolate federated data heterogeneity from baseline detector instability, we employ the Cross-Modal Transformer (CMT). 
CMT provides strong robustness via attention-based multimodal fusion~\cite{cmt} and outperforms state-of-the-art methods on synthetic corruption benchmarks~\cite{MultiCorrupt}.
Across all experiments, detector configuration and hyperparameters remain constant to isolate the impact of federated strategies.

\subsection{Federated Strategies}
The benchmark compares strategies that either improve a shared global model or preserve client-specific parameters for personalization.
Globalization approaches include FedAvg~\cite{mcmahan2017communicationefficientlearningdeepnetworks}, FedDyn~\cite{acar2021feddyn} to limit client drift, and PCGrad~\cite{yu2020pcgrad} to align client gradients. 
Personalization baselines are divided into two categories. Layer-wise methods include FedBN~\cite{li2021fedbn} for personalized batch normalization and FedRep~\cite{collins2021fedrep} for head personalization. Meanwhile, parameter-selection methods include FedSelect~\cite{tamirisa2024fedselect} to personalize the highest-updating parameters and FedMC~\cite{fedmc} for the most influential parameters. All methods are run for 40 rounds with one local epoch.

\subsection{Evaluation}
To contextualize the benchmark, we establish two reference points: isolated own-domain training (a client's baseline without federated collaboration) and centralized training (the expected upper bound of the federated system).
These points contextualize the federated results, which are reported in mean Average Precision (mAP), nuScenes Detection Score (NDS), and Car Average Precision (CAP). 
Given that domains C, D, and E lack 1, 3, and 6 of the 10 target classes, respectively, mAP and NDS may be skewed. 
The mAP and NDS are computed only over classes or attributes that are available in the validation domains.
Consequently, to provide a stable, high-volume single-class comparison, we report CAP, which is the mean over the four nuScenes distance thresholds.
Finally, to account for the imbalance in validation scene counts across domains, all reported averages are weighted by the number of validation scenes.

%% file: sec/5_experiments.tex
\begin{table*}[t]
\centering
\small
\caption{Comparison of FL methods across domains. Non-federated runs are in italics. Performance is reported as mAP / NDS / CAP (Car AP); Avg. denotes the validation-scene-weighted average. Within each family, the best-performing methods are bolded, and the second-best are underlined. For FedCKA, threshold-based masking uses $\tau=0.25$}
\label{tab:fed_results}
\setlength{\tabcolsep}{4pt}
\begin{tabular*}{\linewidth}{@{\extracolsep{\fill}} l c c c c c c @{}}
\toprule
Method & Domain A & Domain B & Domain C & Domain D & Domain E & Avg. \\

\midrule
\textit{Centralized} & \textit{0.61 / 0.66 / 0.84} & \textit{0.61 / 0.68 / 0.85} & \textit{0.72 / 0.67 / 0.83} & \textit{0.60 / 0.64 / 0.85} & \textit{0.58 / 0.58 / 0.84} & \textit{0.65 / 0.66 / 0.84} \\
\textit{Own Domain Only} & \textit{0.58 / 0.64 / 0.81} & \textit{0.47 / 0.51 / 0.75} & \textit{0.59 / 0.62 / 0.76} & \textit{0.14 / 0.24 / 0.47} & \textit{0.01 / 0.08 / 0.04} & \textit{0.52 / 0.57 / 0.74} \\

\midrule
\multicolumn{7}{c}{\textit{Globalization}} \\
\midrule
FedAvg~\cite{mcmahan2017communicationefficientlearningdeepnetworks} & 0.28 / 0.37 / 0.54 & 0.27 / 0.36 / 0.51 & 0.34 / 0.38 / 0.53 & 0.28 / 0.35 / 0.56 & 0.20 / 0.33 / 0.51 & 0.30 / 0.37 / 0.53 \\
FedDyn~\cite{acar2021feddyn} & \textbf{0.38} / \textbf{0.48} / \textbf{0.60} & \textbf{0.34} / \textbf{0.46} / \textbf{0.59} & \textbf{0.42} / \textbf{0.48} / \textbf{0.60} & \underline{0.32} / \textbf{0.43} / \textbf{0.63} & \textbf{0.25} / \textbf{0.35} / \textbf{0.63} & \textbf{0.38} / \textbf{0.47} / \textbf{0.60} \\
PCGrad~\cite{yu2020pcgrad, zhang2024fedgh} & \underline{0.34} / \underline{0.41} / \underline{0.56} & \underline{0.32} / \underline{0.41} / \underline{0.56} & \underline{0.40} / \underline{0.44} / \underline{0.57} & \textbf{0.33} / \underline{0.39} / \underline{0.59} & \underline{0.21} / \underline{0.34} / \underline{0.55} & \underline{0.36} / \underline{0.42} / \underline{0.57} \\
\midrule
\multicolumn{7}{c}{\textit{Personalization}} \\
\midrule
FedRep~\cite{collins2021fedrep}
& 0.33 / 0.43 / 0.58
& 0.28 / 0.40 / 0.54
& 0.37 / 0.44 / 0.57
& 0.28 / 0.41 / 0.59
& 0.21 / 0.33 / \underline{0.56}
& 0.33 / 0.43 / 0.57 \\

FedBN~\cite{li2021fedbn}
& 0.36 / 0.41 / 0.58
& 0.33 / 0.41 / 0.59
& 0.42 / 0.43 / 0.58
& 0.32 / 0.36 / 0.62
& 0.23 / 0.33 / 0.51
& 0.37 / 0.41 / 0.58 \\

FedMC~\cite{fedmc}
& \underline{0.52} / 0.58 / 0.76
& \underline{0.48} / \underline{0.56} / \underline{0.76}
& \underline{0.57} / \underline{0.58} / \underline{0.74}
& \underline{0.37} / \underline{0.47} / \underline{0.73}
& 0.19 / 0.32 / 0.53
& \underline{0.51} / \underline{0.56} / \underline{0.75} \\

FedSelect~\cite{tamirisa2024fedselect}
& \underline{0.52} / \underline{0.59} / \underline{0.77}
& 0.46 / 0.51 / 0.72
& 0.55 / 0.56 / 0.73
& 0.32 / 0.41 / 0.71
& \underline{0.27} / \underline{0.34} / 0.53
& 0.50 / 0.55 / 0.74 \\

FedCKA \textit{(Ours)}
& \textbf{0.59} / \textbf{0.65} / \textbf{0.81}
& \textbf{0.55} / \textbf{0.63} / \textbf{0.81}
& \textbf{0.61} / \textbf{0.64} / \textbf{0.80}
& \textbf{0.53} / \textbf{0.59} / \textbf{0.83}
& \textbf{0.41} / \textbf{0.50} / \textbf{0.82}
& \textbf{0.58} / \textbf{0.63} / \textbf{0.81} \\
\bottomrule
\end{tabular*}
\end{table*}

\begin{figure*}[t]
    \centering
    \vspace{-1em}
    \includegraphics[width=\textwidth]{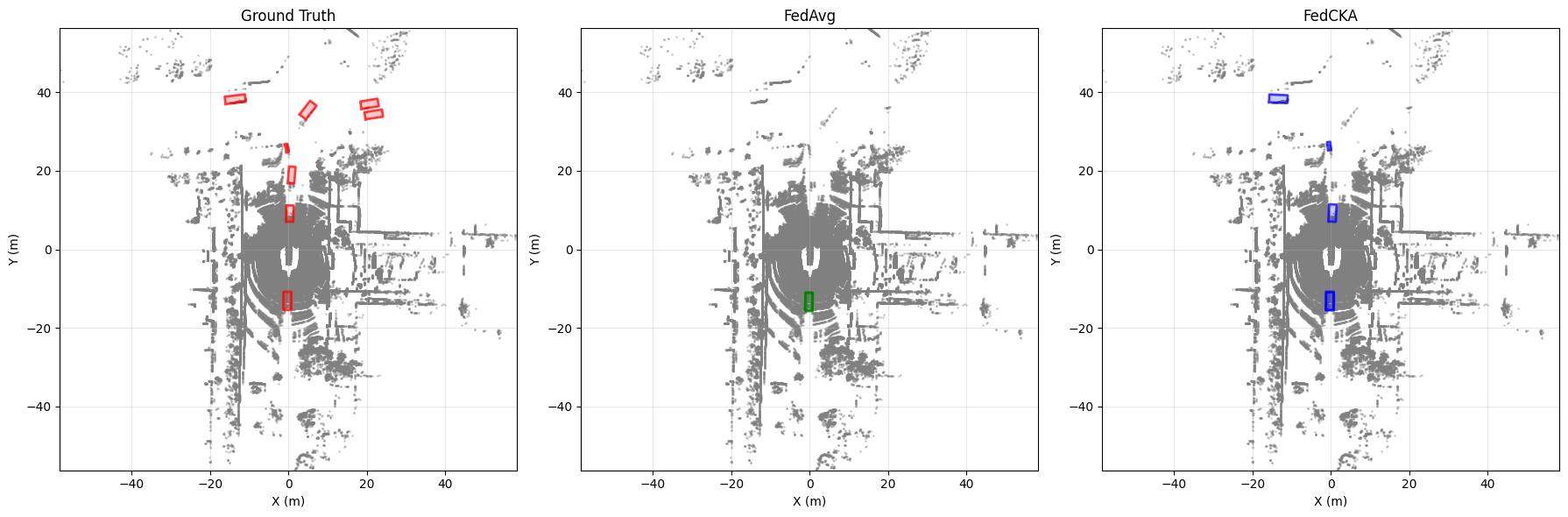}
    \caption{Qualitative comparison of 3D object detection performance under severe domain shift (Singapore night rain) in BEV view. On the left are the Ground-Truth bounding boxes. In the middle are the predictions from the baseline FedAvg model, illustrating missed detections. On the right, the predictions from the proposed FedCKA framework show successfully recovered detections. 
    }
    \label{fig:qual_bev_three}
\end{figure*}

\section{Experiments} \label{sec:experiments}
Table~\ref{tab:fed_results} compares FedCKA with the federated and non-federated references defined above. Table~\ref{tab:fed_results} demonstrates that our approach outperforms all federated baselines and most closely approaches the centralized reference, achieving a validation-scene-weighted NDS of 0.63, compared to 0.55 for FedSelect and 0.56 for FedMC.
Qualitative results for the most shifted domain, Singapore night rain, are also shown in Fig.~\ref{fig:qual_bev_three}.

\subsection{Globalization versus Personalization}
Traditional globalization methods do not produce a consensus model that rivals centralized performance. Static-layer personalization strategies, FedBN and FedRep, which personalize the batch normalization (BN) layers or the head of each client, perform comparably to globalization methods. However, even with this targeted adaptation, these models fall short. These BN and head layers may not have sufficient power to move the client models towards their local optima, as they have fewer personal parameters than the parameter-level personalization strategies and do not influence the crucial intermediate cross-modal merging stages.

Among the established baselines, the parameter-level personalization strategies of FedSelect and FedMC achieve the highest performance. 
However, the proposed FedCKA method achieves the highest performance across the three metrics (mAP, NDS, CAP), suggesting that adaptive layer-wise personalization better accommodates domain discrepancies than fixed or parameter-magnitude-based personalization.

Ultimately, the difficulty of achieving a global consensus model across these clients highlights the severity of the domain gap. Because globalization forces different domains toward a single shared optimum, the resulting model remains suboptimal. Conversely, the success of the parameter-level baselines and FedCKA demonstrates that, by allowing a sufficiently large or important group of parameters to adapt to a domain, client models can move closer to their local optima.

\begin{table}[t]
\centering
\small
\caption{Cross-domain generalization matrices for client models \textbf{M} A--E. Each matrix reports NDS performance across validation domains \textbf{D} A--E. The diagonal shows performance when the client is evaluated on its own local domain, while the off-diagonal entries show cross-domain performance.}
\label{tab:cross_domain_all}

\begin{subtable}{\linewidth}
\centering
\caption{Own Domain Only}
\label{tab:NoComm_cross_domain}
\makebox[\linewidth][c]{%
\begin{NiceTabular}{w{c}{0.7cm} *{5}{w{c}{0.8cm}}}[hvlines, cell-space-limits=2pt]
 & \textbf{D-A} & \textbf{D-B} & \textbf{D-C} & \textbf{D-D} & \textbf{D-E} \\
\textbf{M-A} & \Block[fill=blue!74]{}{0.64} & \Block[fill=blue!69]{}{0.60} & \Block[fill=blue!53]{}{0.46} & \Block[fill=blue!38]{}{0.33} & \Block[fill=blue!17]{}{0.15} \\
\textbf{M-B} & \Block[fill=blue!46]{}{0.40} & \Block[fill=blue!59]{}{0.51} & \Block[fill=blue!32]{}{0.28} & \Block[fill=blue!24]{}{0.21} & \Block[fill=blue!14]{}{0.12} \\
\textbf{M-C} & \Block[fill=blue!55]{}{0.48} & \Block[fill=blue!48]{}{0.42} & \Block[fill=blue!72]{}{0.62} & \Block[fill=blue!48]{}{0.42} & \Block[fill=blue!25]{}{0.22} \\
\textbf{M-D} & \Block[fill=blue!30]{}{0.26} & \Block[fill=blue!27]{}{0.23} & \Block[fill=blue!33]{}{0.29} & \Block[fill=blue!28]{}{0.24} & \Block[fill=blue!22]{}{0.19} \\
\textbf{M-E} & \Block[fill=blue!9]{}{0.08} & \Block[fill=blue!9]{}{0.08} & \Block[fill=blue!9]{}{0.08} & \Block[fill=blue!7]{}{0.06} & \Block[fill=blue!9]{}{0.08} \\
\end{NiceTabular}%
}
\end{subtable}

\vspace{2mm}

\begin{subtable}{\linewidth}
\centering
\caption{FedSelect}
\label{tab:fedselect_cross_domain}
\makebox[\linewidth][c]{%
\begin{NiceTabular}{w{c}{0.7cm} *{5}{w{c}{0.8cm}}}[hvlines, cell-space-limits=2pt]
 & \textbf{D-A} & \textbf{D-B} & \textbf{D-C} & \textbf{D-D} & \textbf{D-E} \\
\textbf{M-A} & \Block[fill=blue!68]{}{0.59} & \Block[fill=blue!62]{}{0.54} & \Block[fill=blue!59]{}{0.51} & \Block[fill=blue!51]{}{0.44} & \Block[fill=blue!44]{}{0.38} \\
\textbf{M-B} & \Block[fill=blue!52]{}{0.45} & \Block[fill=blue!59]{}{0.51} & \Block[fill=blue!45]{}{0.39} & \Block[fill=blue!44]{}{0.38} & \Block[fill=blue!38]{}{0.33} \\
\textbf{M-C} & \Block[fill=blue!47]{}{0.41} & \Block[fill=blue!42]{}{0.36} & \Block[fill=blue!65]{}{0.56} & \Block[fill=blue!57]{}{0.49} & \Block[fill=blue!54]{}{0.47} \\
\textbf{M-D} & \Block[fill=blue!32]{}{0.28} & \Block[fill=blue!28]{}{0.24} & \Block[fill=blue!40]{}{0.35} & \Block[fill=blue!47]{}{0.41} & \Block[fill=blue!38]{}{0.33} \\
\textbf{M-E} & \Block[fill=blue!31]{}{0.27} & \Block[fill=blue!28]{}{0.24} & \Block[fill=blue!37]{}{0.32} & \Block[fill=blue!39]{}{0.34} & \Block[fill=blue!39]{}{0.34} \\
\end{NiceTabular}%
}
\end{subtable}
  
\vspace{2mm}

\begin{subtable}{\linewidth}
\centering
\caption{FedCKA}
\label{tab:FedCKA_cross_domain}
\makebox[\linewidth][c]{%
\begin{NiceTabular}{w{c}{0.7cm} *{5}{w{c}{0.8cm}}}[hvlines, cell-space-limits=2pt]
 & \textbf{D-A} & \textbf{D-B} & \textbf{D-C} & \textbf{D-D} & \textbf{D-E} \\
\textbf{M-A} & \Block[fill=blue!75]{}{0.65} & \Block[fill=blue!69]{}{0.60} & \Block[fill=blue!68]{}{0.59} & \Block[fill=blue!60]{}{0.52} & \Block[fill=blue!54]{}{0.47} \\
\textbf{M-B} & \Block[fill=blue!67]{}{0.58} & \Block[fill=blue!73]{}{0.63} & \Block[fill=blue!65]{}{0.56} & \Block[fill=blue!57]{}{0.49} & \Block[fill=blue!52]{}{0.45} \\
\textbf{M-C} & \Block[fill=blue!65]{}{0.56} & \Block[fill=blue!63]{}{0.55} & \Block[fill=blue!74]{}{0.64} & \Block[fill=blue!61]{}{0.53} & \Block[fill=blue!55]{}{0.48} \\
\textbf{M-D} & \Block[fill=blue!52]{}{0.45} & \Block[fill=blue!51]{}{0.44} & \Block[fill=blue!61]{}{0.53} & \Block[fill=blue!68]{}{0.59} & \Block[fill=blue!61]{}{0.53} \\
\textbf{M-E} & \Block[fill=blue!54]{}{0.47} & \Block[fill=blue!54]{}{0.47} & \Block[fill=blue!60]{}{0.52} & \Block[fill=blue!52]{}{0.45} & \Block[fill=blue!58]{}{0.50} \\
\end{NiceTabular}%
}
\end{subtable}

\end{table}
\subsection{Cross-Domain Generalization}
Cross-domain generalization is a key test of robustness in a domain-heterogeneous setting, revealing the limitations of both isolated training and standard aggregation, as detailed in Table~\ref{tab:cross_domain_all}. 
Table~\ref{tab:NoComm_cross_domain} shows that although isolated training yields rich local models for domains A and C, their cross-domain transfer remains poor.
For instance, Model A drops to an NDS of 0.15 on domain E, indicating that strong own-domain performance does not necessarily translate into general cross-domain knowledge.
This underscores that the benchmark captures meaningful domain heterogeneity rather than only differences in local dataset size.

Conversely, global aggregation with FedAvg, FedDyn, or PCGrad, as shown in Table~\ref{tab:fed_results}, sacrifices local performance on the richer domains A, B, and C, while maintaining stable but limited cross-domain performance. Because globalization forces all clients to share a single global model, a cross-domain evaluation matrix for these methods yields identical rows for every client, indicating stable yet consistently suboptimal generalization across all domains.

Personalized federated methods aim to mitigate this trade-off by retaining client-specific components while still benefiting from collaborative training.
As shown in Table~\ref{tab:fedselect_cross_domain}, FedSelect improves cross-domain transfer on average over isolated training, confirming that personalized models can still benefit from shared knowledge across clients.

Compared with FedSelect, Table~\ref{tab:FedCKA_cross_domain} shows that FedCKA, which aggregates only layers with high representational similarity, boosts not only cross-domain performance but also own-domain performance (shown on the diagonal), surpassing all federated methods previously evaluated in Table~\ref{tab:fed_results}. 
It is the only federated method where all five clients improve over training only on their own local datasets, and the cross-domain transfer achieves the highest average off-diagonal NDS. 
This suggests that FedCKA enables beneficial knowledge transfer across clients while preserving domain-specific representations through personalization.

\subsection{Masking Strategy Ablation} 
Both proposed personalization strategies, detailed in Section~\ref{subsec:personalization_strats_method}, depend on either a personalization ratio ($\rho$) or a similarity threshold ($\tau$). At the upper bound ($\rho=1$ or $\tau=1$), the models are entirely personalized (except those with mathematically perfect representational alignment), reducing to the ``Own Domain Only'' baseline in Table~\ref{tab:fed_results}; at the lower bound ($\rho=0$ or $\tau=0$), the system defaults to FedAvg aggregation. Table~\ref{tab:cka_ablation_summary} summarizes the impact of varying these parameters. 
Although thresholding is monotonic for a fixed set of CKA scores, each $\tau$ induces a different training trajectory and permits later layer re-integration, so the personalization ratio need not vary monotonically with $\tau$.
The $\rho=0.40$ final personalization target is additionally used as an ablation reference to compare the fixed personalization ratio used by FedSelect and FedMC.
For other detectors or applications, the masking hyperparameter can be selected according to the desired trade-off between global alignment and client-specific adaptation. A small set of values spanning this range can first be evaluated, after which the search can be refined toward the better-performing region.
The resulting masks remain stable across repeated runs and consistently isolate the same structural components, suggesting that FedCKA reliably targets domain-sensitive layers rather than collapsing into arbitrary or fully local personalization.

\begin{table}[t]
    \centering
    \caption{FedCKA ablations for percentage- and threshold-based masking in Eqs.~\ref{eq:percentage_masking} and \ref{eq:threshold_masking}, reporting average NDS and final client personalization ratio. The two masking strategies are mutually exclusive in these experiments: percentage-based masking uses \(\rho\), while threshold-based masking uses \(\tau\).}
    \label{tab:cka_ablation_summary}
    \footnotesize
    \setlength{\tabcolsep}{5pt}
    \begin{tabular}{@{}lcc|lcc@{}}
    \toprule
    Setting & NDS & Mask & Setting & NDS & Mask \\
    \midrule
    $\rho=0.01$ & 0.551 & 1.2\%  & $\tau=0.05$ & 0.598 & 0.7\% \\
    $\rho=0.10$ & 0.489 & 10.7\% & $\tau=0.25$ & 0.635 & 4.7\% \\
    $\rho=0.40$ & 0.516 & 41.3\% & $\tau=0.50$ & 0.600 & 7.8\% \\
                &       &         & $\tau=0.75$ & 0.596 & 4.4\% \\
    \bottomrule
    \end{tabular}
\end{table}

\subsection{Communication and Computation Footprint}
To analyze the trade-off between parameter sharing and local specialization, we estimate client-to-server upload payloads based on the average fraction of shared parameters. We model mask metadata as bit-packed binary masks: layer-level for FedBN, FedRep, and FedCKA, and parameter-level for FedMC and FedSelect.
This first-order estimate excludes protocol overhead, latency, compression, optimizer states, and server-to-client broadcasts.

\begin{table}[t]
\centering
\caption{Cumulative client-to-server upload over $40$ federated rounds with $5$ clients and a $125$~MB base model. Mask metadata is reported per active masking round across all clients; total payload is cumulative over training. Avg. pers. is the client- and round-averaged fraction withheld from aggregation; thus, the $40\%$ target used by FedSelect and FedMC corresponds to $20\%$ on average. Raw-data centralization is included for reference.}
\label{tab:communication}
\footnotesize
\setlength{\tabcolsep}{2pt}
\begin{tabular}{lccc}
\toprule
Method & Avg. pers. & Mask metadata & Total upload payload \\
\midrule
FedAvg    & 0.0\%  & --        & 25.00 GB \\
FedDyn    & 0.0\%  & --        & 25.00 GB \\
PCGrad    & 0.0\%  & --        & 25.00 GB \\
FedBN     & 0.2\%  & 0.25 KB   & 24.95 GB \\
FedRep    & 6.0\%  & 0.25 KB   & 23.50 GB \\
FedCKA    & 3.5\% & 0.25 KB   & 24.12 GB \\
FedMC     & 20.0\% & 20.43 MB  & 20.82 GB \\
FedSelect & 20.0\% & 20.43 MB  & 20.82 GB \\
\midrule
Raw-data centralization$^\dagger$ & -- & -- & $\sim$292 GB \\
\bottomrule
\end{tabular}

\vspace{1mm}
\parbox{0.98\linewidth}{\scriptsize
$^\dagger$ One-time upload of the camera and LiDAR sensor data used by the five clients.
}
\end{table}

As shown in Table~\ref{tab:communication}, FedCKA retains an upload payload comparable to full-model aggregation, reinforcing that its primary objective is dynamic personalization rather than communication compression. By using representation similarity, FedCKA flexibly decides which layers to share and which to retain locally, allowing clients to personalize different structural subsets with negligible metadata overhead. 
In contrast, parameter-level masking methods reduce the transmitted model payload more substantially in our experiments because they use higher personalized parameter fractions; at matched personalization ratios, FedCKA incurs lower masking metadata because it operates at the layer level.

The additional computational cost of FedCKA comes from the client-side forward passes required to compute layer-wise CKA similarities. In our setup, this adds approximately $14{,}000$ seconds of computation over the full training procedure, corresponding to $2.7\%$ of the total training time, with a peak VRAM usage of 6 GB during the CKA computation.

\subsection{Robustness Analysis}

To assess run-to-run variability, we repeat the main non-federated and personalized federated experiments three times. Table~\ref{tab:method_comparison} reports the mean and standard deviation across these runs. FedCKA achieves an average NDS of $0.635 \pm 0.008$ and CAP of $0.810 \pm 0.014$, remaining the strongest personalized federated method across the repeated experiments. Although some variation is present between runs, the observed performance margin over the competing personalized methods is substantially larger than the corresponding run-to-run standard deviations. These results indicate that FedCKA's performance gains are not attributable to a single favorable training run.

\begin{table}[t]
    \centering
    \caption{Comparison of non-federated baselines and personalized federated learning methods. Results are reported as mean $\pm$ standard deviation over three runs.}
    \label{tab:method_comparison}
    \footnotesize
    \setlength{\tabcolsep}{7pt}
    \begin{tabular}{@{}lcc@{}}
    \toprule
    Method & NDS & CAP \\
    \midrule
    Centralized
        & $0.658 \pm 0.014$
        & $0.830 \pm 0.014$ \\
    Own Domain Only
        & $0.567 \pm 0.004$
        & $0.739 \pm 0.002$ \\
    FedMC
        & $0.559 \pm 0.006$
        & $0.744 \pm 0.003$ \\
    FedSelect
        & $0.545 \pm 0.004$
        & $0.736 \pm 0.008$ \\
    FedCKA
        & $0.635 \pm 0.008$
        & $0.810 \pm 0.014$ \\
    \bottomrule
    \end{tabular}
\end{table}

We additionally evaluate the robustness of the CKA-based masking decisions to the number of local samples used to estimate representation similarity. Across three independent runs, using $10$ randomly sampled inputs produced the same layer-selection decisions as using $100$ samples. We therefore use $10$ samples throughout the experiments. Moreover, because masking decisions are re-evaluated every round, layers erroneously personalized by noisy CKA estimates can revert to global aggregation for the main method. This choice is specific to the evaluated dataset and detector and should not be interpreted as a universally optimal sample count.

%% file: sec/6_discussion.tex
\section{Discussion} \label{sec:discussion}
The results on this problem-specific benchmark indicate that, for federated, data-heterogeneous 3D object detection, the global aggregation methods underperform relative to certain personalization methods.
Although manually chosen static layer personalization provides a baseline improvement over FedAvg, directly inferring personalization masks from the client models further boosts performance.
Compared with magnitude- or gradient-based personalization, the results suggest that representation similarity is a stronger criterion for selecting personalized layer or parameter subsets, improving both out-of-domain and smaller-domain performance.

FedCKA reduces the need to hand-pick specific layer types or impose fixed personalization ratios, since the masking decision is instead based on measured representational similarity. Furthermore, FedCKA does not assume that same-index parameters serve the same function across clients, since it compares layer activations rather than parameter-space differences. 

Inspection of the learned masks suggests the architectural sources of domain sensitivity. Normalization layers are frequently personalized, reflecting their sensitivity to domain-specific feature scaling driven by weather, illumination, and geography. Similarly, personalization in deeper backbone, neck, and attention layers indicates that intermediate cross-modal representations, particularly the fusion weights of camera and LiDAR inputs, can be domain-sensitive. 
The task head remains largely shared, which may help preserve knowledge of classes absent from specific local domains and suggests that object-level prediction semantics are transferable, while the internal representations may adjust due to various shifts. 
This indicates that FedCKA does not create independent local models but instead preserves a common detection-task structure.


%% file: sec/7_limitations.tex
\section{Limitations} \label{sec:limitations}
The primary limitation of this study is that the evaluation is restricted to a single detector, dataset, and domain split. The domain split is also highly imbalanced, meaning that the observed gains for smaller domains may depend on the specific scene distribution and validation composition. 

A second limitation is that the framework assumes supervised local training, which requires the availability of reliable client-side labels or a labeling pipeline. This assumption is restrictive because newly collected IV data does not naturally include high-quality 3D bounding boxes. Although offline autolabelling on idle vehicles could provide pseudo-labels, this work does not evaluate how pseudo-label noise would affect federated training.

Another limitation is that although the CKA computation is relatively lightweight compared to model training, it still incurs additional computational costs for the client. It also requires flattening intermediate representations into two-dimensional matrices, which may simplify the original positional information.
Although this simplification potentially limits the captured positional dependencies, our results indicate that the flattened representations remain effective for layer-wise similarity assessment in the evaluated setting. 

%% file: sec/8_conclusion.tex
\section{Conclusion and Future Work} \label{sec:future_work}
This work studied federated learning for data-heterogeneous 3D object detection under shifts in weather, time, and location, showing that adaptive partial personalization can outperform a single consensus model on this benchmark.
The findings suggest that representation similarity is a useful criterion for balancing this globalization and local adaptation. For real-world robotic perception in IV deployment, cross-domain robustness and local domain adaptation are both vital, and these findings may help clarify which model layers are best suited for local adaptation.

Future work should evaluate whether the observed benefits hold across different detectors, datasets, domain definitions, and types of heterogeneity. Methodologically, FedCKA could be extended with adaptive thresholds, temporally smoothed masks, unsupervised domain clustering based on CKA scores, and further analysis of mask stability across domain batches and batch sizes. Furthermore, a separate direction is to evaluate whether masking reduces global drift, and whether FedCKA remains effective under pseudo-labeled local training from offline autolabelling pipelines.

Finally, FedCKA does not, by itself, provide formal privacy guarantees. Combining CKA-guided masking with differential privacy is therefore an important direction for privacy-preserving deployment.
Overall, this work shows that representation similarity can provide a principled and empirically effective basis for deciding what should remain shared and what should become client-specific in federated 3D-OD.

%% file: sec/9_acknowledgements.tex
\section{Acknowledgements}
The Cynergy4MIE project is supported by the Chips Joint Undertaking and its members, including the top-up funding by National Authorities under Grant Agreement No 101140226. 

Research reported in this work was partially facilitated by computational resources and support of the Delft AI Cluster (DAIC) at TU Delft (RRID: SCR\_025091), but remains the sole responsibility of the authors, not the DAIC team.

The authors acknowledge the use of Google Gemini, ChatGPT, and Grammarly to assist in the preparation of this manuscript. These were utilized throughout the work for structural editing, rephrasing, checking internal consistency, and visually enhancing Figure 2. The authors iteratively guided, carefully reviewed, and refined all AI-assisted edits and take full responsibility for the final content and scientific integrity of the work.

%% file: main.bib
@String(CVPR= {IEEE Conf. Comput. Vis. Pattern Recog.})

@String(ICCV= {Int. Conf. Comput. Vis.})

@String(ECCV= {Eur. Conf. Comput. Vis.})

@String(ICLR = {Int. Conf. Learn. Represent.})

@String(CVPR  = {CVPR})

@String(ICCV  = {ICCV})

@String(ECCV  = {ECCV})

@String(ICLR  = {ICLR})

@IEEEtranBSTCTL{IEEEexample:BSTcontrol,
  CTLuse_url = "no",
  CTLuse_forced_etal       = "yes",
  CTLmax_names_forced_etal = "3",
  CTLnames_show_etal       = "1"
}

@InProceedings{Caesar2020nuScenes,
author = {Caesar, Holger and Bankiti, Varun and Lang, Alex H. and Vora, Sourabh and Liong, Venice Erin and Xu, Qiang and Krishnan, Anush and Pan, Yu and Baldan, Giancarlo and Beijbom, Oscar},
title = {{nuScenes}: A Multimodal Dataset for Autonomous Driving},
booktitle = {Proc. CVPR},
year = {2020},
pages = {11621-11631}
}

@InProceedings{kornblith2019similarity_CKA,
  author    = {Kornblith, Simon and Norouzi, Mohammad and Lee, Honglak and Hinton, Geoffrey},
  title     = {Similarity of Neural Network Representations Revisited},
  booktitle = {Proc. ICML},
  pages     = {3519--3529},
  year      = {2019},
}

@InProceedings{gretton2005measuring_HSIC,
  author    = {Gretton, Arthur and Bousquet, Olivier and Smola, Alex and Sch{\"o}lkopf, Bernhard},
  title     = {Measuring Statistical Dependence with {Hilbert-Schmidt} Norms},
  booktitle = {Proc. ALT},
  pages     = {63--78},
  year      = {2005},
}

@InProceedings{li2020fedprox,
  author    = {Li, Tian and Sahu, Anit Kumar and Zaheer, Manzil and Sanjabi, Maziar and Talwalkar, Ameet and Smith, Virginia},
  title     = {Federated Optimization in Heterogeneous Networks},
  booktitle = {Proc. MLSys},
  pages     = {429--450},
  year      = {2020}
}

@InProceedings{acar2021feddyn,
  author    = {Acar, Durmus Alp Emre and Zhao, Yue and Matas Navarro, Ramon and Mattina, Matthew and Whatmough, Paul N. and Saligrama, Venkatesh},
  title     = {Federated Learning Based on Dynamic Regularization},
  booktitle = {Proc. ICLR},
  year      = {2021}
}

@InProceedings{karimireddy2020scaffold,
  author    = {Karimireddy, Sai Praneeth and Kale, Satyen and Mohri, Mehryar and Reddi, Sashank and Stich, Sebastian U. and Suresh, Ananda Theertha},
  title     = {{SCAFFOLD}: Stochastic Controlled Averaging for Federated Learning},
  booktitle = {Proc. ICML},
  pages     = {5132--5143},
  year      = {2020},
}

@InProceedings{ilharco2023task_arithmetic,
  author    = {Ilharco, Gabriel and Ribeiro, Marco Tulio and Wortsman, Mitchell and Gururangan, Suchin and Schmidt, Ludwig and Hajishirzi, Hannaneh and Farhadi, Ali},
  title     = {Editing Models with Task Arithmetic},
  booktitle = {Proc. ICLR},
  year      = {2023}
}

@InProceedings{yu2020pcgrad,
  author    = {Yu, Tianhe and Kumar, Saurabh and Gupta, Abhishek and Levine, Sergey and Hausman, Karol and Finn, Chelsea},
  title     = {Gradient Surgery for Multi-Task Learning},
  booktitle = {Proc. NeurIPS},
  pages     = {5824--5836},
  year      = {2020}
}

@InProceedings{nguyen2025fedomg,
  author    = {Nguyen, Trong Binh and Nguyen, Duong Minh and Park, Jinsun and Pham, Viet Quoc and Hwang, Won-Joo},
  title     = {Federated Domain Generalization with Data-Free On-Server Matching Gradient},
  booktitle = {Proc. ICLR},
  year      = {2025}
}

@InProceedings{li2021fedbn,
  author    = {Li, Xiaoxiao and Jiang, Meirui and Zhang, Xiaofei and Kamp, Michael and Dou, Qi},
  title     = {{FedBN}: Federated Learning on {non-IID} Features via Local Batch Normalization},
  booktitle = {Proc. ICLR},
  year      = {2021}
}

@InProceedings{collins2021fedrep,
  author    = {Collins, Liam and Hassani, Hamed and Mokhtari, Aryan and Shakkottai, Sanjay},
  title     = {Exploiting Shared Representations for Personalized Federated Learning},
  booktitle = {Proc. ICML},
  pages     = {2089--2099},
  year      = {2021},
}

@InProceedings{tamirisa2024fedselect,
  author    = {Tamirisa, Rishub and Xie, Chulin and Bao, Wenxuan and Zhou, Andy and Arel, Ron and Shamsian, Aviv},
  title     = {{FedSelect}: Personalized Federated Learning with Customized Selection of Parameters for Fine-Tuning},
  booktitle = {Proc. CVPR},
  pages     = {29485--29494},
  year      = {2024},
}

@article{zhang2024fedgh,
  title   = {Tackling the {non-IID} Issue in Heterogeneous Federated Learning by Gradient Harmonization},
  author  = {Zhang, Xinyu and Sun, Weiyu and Chen, Ying},
  journal = {IEEE Signal Processing Letters},
  volume  = {31},
  pages   = {2595--2599},
  year    = {2024},
  doi     = {10.1109/LSP.2024.3430042}
}

@inproceedings{fedmc,
author = {Gao, Yingying and He, Xin and Chen, Yan},
title = {Personalized Federated Learning Algorithm Based on Information Content Model Customization},
year = {2025},
isbn = {9798400713873},
url = {https://doi.org/10.1145/3745533.3745668},
doi = {10.1145/3745533.3745668},
booktitle = {Proc. CAMMIC},
pages = {834-838},
numpages = {5},
location = {
}
}

@article{fedpillarnet,
title = {{FedPillarNet}: Unifying personalized and global features for federated {3D LiDAR} object detection},
journal = {Journal of Systems Architecture},
pages = {103584},
volume = {168},
year = {2025},
issn = {1383-7621},
doi = {https://doi.org/10.1016/j.sysarc.2025.103584},
url = {https://www.sciencedirect.com/science/article/pii/S1383762125002565},
author = {Boyang Li and Siheng Ren and Shuai Zhao and Mingyue Cui and Kai Huang}
}

@inproceedings{yosinski2014transferable_diffpartsdifffeatures,
author = {Yosinski, Jason and Clune, Jeff and Bengio, Yoshua and Lipson, Hod},
title = {How transferable are features in deep neural networks?},
year = {2014},
booktitle = {Proc. NeurIPS},
pages = {3320-3328},
numpages = {9},
location = {Montreal, Canada},
}

@inproceedings{hsu2020federated_noniiddegradation,
  title     = {Federated Visual Classification with Real-World Data Distribution},
  author    = {Hsu, Tzu-Ming Harry and Qi, Hang and Brown, Matthew},
  booktitle = {Proc. ECCV},
  pages     = {76--92},
  year      = {2020},
  doi       = {10.1007/978-3-030-58607-2_5}
}

@ARTICLE{li2025federateddg_crossdomaingeneralization,
  author={Li, Ying and Wang, Xingwei and Zeng, Rongfei and Kumar Donta, Praveen and Murturi, Ilir and Huang, Min and Dustdar, Schahram},
  journal={Proceedings of the IEEE}, 
  title={Federated Domain Generalization: A Survey}, 
  year={2025},
  volume={113},
  number={4},
  pages={370-410},
  doi={10.1109/JPROC.2025.3596173}}

@inproceedings{singh2020model,
author = {Singh, Sidak Pal and Jaggi, Martin},
title = {Model fusion via optimal transport},
year = {2020},
isbn = {9781713829546},
booktitle = {Proc. NeurIPS},
articleno = {1849},
numpages = {11},
location = {Vancouver, BC, Canada},
pages={1849}
}

@InProceedings{raghu2021visiontransformers,
  author    = {Raghu, Maithra and Unterthiner, Thomas and Kornblith, Simon and Zhang, Chiyuan and Dosovitskiy, Alexey},
  title     = {Do Vision Transformers See Like Convolutional Neural Networks?},
  booktitle = {Proc. NeurIPS},
  pages     = {12116--12128},
  year      = {2021}
}

@article{Cai_Wu_2025,
title = {Three-dimensional object detection for autonomous driving via deep learning: A review},
journal = {Engineering Applications of Artificial Intelligence},
volume = {161},
pages = {112238},
year = {2025},
issn = {0952-1976},
doi = {https://doi.org/10.1016/j.engappai.2025.112238},
url = {https://www.sciencedirect.com/science/article/pii/S0952197625022468},
author = {Jiaqi Cai and Yiquan Wu}
}

@Article{s25092794,
  author  = {Wang, Hai and Liu, Junhao and Dong, Haoran and Shao, Zheng},
  title   = {A survey of the multi-sensor fusion object detection task in autonomous driving},
  journal = {Sensors},
  volume  = {25},
  number  = {9},
  pages   = {2794},
  year    = {2025}
}

@INPROCEEDINGS{MultiCorrupt,
  author={Beemelmanns, Till and Zhang, Quan and Geller, Christian and Eckstein, Lutz},
  booktitle={Proc. IV}, 
  title={{MultiCorrupt}: A Multi-Modal Robustness Dataset and Benchmark of {LiDAR}-Camera Fusion for {{3D}} Object Detection}, 
  year={2024},
  volume={},
  number={},
  pages={3255-3261},
  doi={10.1109/IV55156.2024.10588664}}

@InProceedings{nuScenesC,
    author    = {Dong, Yinpeng and Kang, Caixin and Zhang, Jinlai and Zhu, Zijian and Wang, Yikai and Yang, Xiao and Su, Hang and Wei, Xingxing and Zhu, Jun},
    title     = {Benchmarking Robustness of {3D} Object Detection to Common Corruptions},
    booktitle = {Proc. CVPR},
    year      = {2023},
    pages     = {1022-1032}
}

@Article{fedm2former,
AUTHOR = {Zhao, Liang and Li, Xuan and Jia, Xin and Fu, Lulu},
TITLE = {Personalized {FedM2former}: An Innovative Approach Towards Federated Multi-Modal {3D} Object Detection for Autonomous Driving},
JOURNAL = {Processes},
VOLUME = {13},
YEAR = {2025},
NUMBER = {2},
ARTICLE-NUMBER = {449},
pages={449},
URL = {https://www.mdpi.com/2227-9717/13/2/449},
ISSN = {2227-9717},
DOI = {10.3390/pr13020449}
}

@InProceedings{cmt,
    author    = {Yan, Junjie and Liu, Yingfei and Sun, Jianjian and Jia, Fan and Li, Shuailin and Wang, Tiancai and Zhang, Xiangyu},
    title     = {Cross Modal Transformer: Towards Fast and Robust {3D} Object Detection},
    booktitle = {Proc. ICCV},
    year      = {2023},
    pages     = {18268-18278}
}

@InProceedings{mcmahan2017communicationefficientlearningdeepnetworks,
  title = 	 {Communication Efficient Learning of Deep Networks from Decentralized Data},
  author = 	 {McMahan, Brendan and Moore, Eider and Ramage, Daniel and Hampson, Seth and Arcas, Blaise Ag{ü}era y},
  booktitle = 	 {Proc. AISTATS},
  pages = 	 {1273--1282},
  year = 	 {2017},
  url = 	 {https://proceedings.mlr.press/v54/mcmahan17a.html}
}

@ARTICLE{zhan,
  author={Pei, Jiaming and Liu, Wenxuan and Li, Jinhai and Wang, Lukun and Liu, Chao},
  journal={IEEE Transactions on Consumer Electronics}, 
  title={A Review of Federated Learning Methods in Heterogeneous Scenarios}, 
  year={2024},
  volume={70},
  number={3},
  pages={5983-5999},
  doi={10.1109/TCE.2024.3385440}
}

@inproceedings{strubell-etal-2019-energy,
    title = "Energy and Policy Considerations for Deep Learning in {NLP}",
    author = "Strubell, Emma  and
      Ganesh, Ananya  and
      McCallum, Andrew",
    booktitle = "Proc. ACL",
    year = "2019",
    url = "https://aclanthology.org/P19-1355/",
    doi = "10.18653/v1/P19-1355",
    pages = "3645--3650",
}

@ARTICLE{10509812,
  author={Liu, Mingyu and Yurtsever, Ekim and Fossaert, Jonathan and Zhou, Xingcheng and Zimmer, Walter and Cui, Yuning and Zagar, Bare Luka and Knoll, Alois C.},
  journal={IEEE Transactions on Intelligent Vehicles}, 
  title={A Survey on Autonomous Driving Datasets: Statistics, Annotation Quality, and a Future Outlook}, 
  year={2024},
  volume={9},
  number={11},
  pages={7138-7164},
  doi={10.1109/TIV.2024.3394735}
}
